\newif\ifANONYMOUS
\ANONYMOUSfalse

\documentclass[conference]{IEEEtran}

\newif\ifDEBUG
\DEBUGfalse

\newif\ifAPPENDIX
\APPENDIXtrue
\APPENDIXfalse

\IEEEoverridecommandlockouts
\usepackage{cite}
\usepackage{graphicx}
\usepackage{textcomp}
\usepackage{xcolor}
\usepackage{listings}
\usepackage{array}
\usepackage{hhline}
\usepackage{multirow}
\usepackage{booktabs}
\usepackage{nicematrix}
\usepackage{colortbl}
\usepackage{algorithm}
\usepackage{algpseudocode}
\usepackage{caption}

\definecolor{lightcyan}{RGB}{225, 245, 254}

\newcolumntype{C}[1]{>{\centering\arraybackslash}p{#1}}

\def\BibTeX{{\rm B\kern-.05em{\sc i\kern-.025em b}\kern-.08em
    T\kern-.1667em\lower.7ex\hbox{E}\kern-.125emX}}

\usepackage[normalem]{ulem} 
\usepackage{xspace}
\usepackage{multirow} 
\usepackage{balance} 
\usepackage{fancyvrb} 
\usepackage{caption}
\usepackage{booktabs} 

\usepackage{enumitem}
\setlist[itemize]{leftmargin=*,noitemsep,topsep=0pt}
\setlist[enumerate]{leftmargin=*}

\usepackage{etoolbox}
\makeatletter
\patchcmd{\@makecaption}
	{\scshape}
	{}
	{}
	{}
\makeatletter
\patchcmd{\@makecaption}
	{\\}
	{.\ }
	{}
	{}
\makeatother
\def\tablename{Table}

\newcommand{\eg}{\textit{e.g.,}\xspace}
\newcommand{\etal}{\textit{et al.}\xspace}

\usepackage{mdframed}
\mdfsetup{skipabove=0.5\topskip,skipbelow=0.5\topskip,align=center}

\usepackage{amsthm}
\newtheorem{thm}{Theorem}

\DeclareMathSymbol{\mlq}{\mathord}{operators}{``}
\DeclareMathSymbol{\mrq}{\mathord}{operators}{`'}

\newif\ifSAVESPACE
\SAVESPACEfalse

\ifSAVESPACE

\else

\fi

\usepackage{todonotes}
\usepackage{soul}
\usepackage{fontawesome}
\usepackage{ragged2e}

\ifDEBUG
    \newcommand{\AH}[1]{\todo[color=cyan,inline]{AH:#1}}
    \newcommand{\AM}[1]{\todo[color=red,inline]{Machiry:#1}}
    \newcommand{\JD}[1]{\todo[color=yellow,inline]{JD:#1}}
    \newcommand{\TL}[1]{\todo[color=green,inline]{SA:#1}}
    \newcommand{\PA}[1]{\todo[color=orange,inline]{PA:#1}}
    
    \newcommand{\KR}[1]{\todo[color=yellow,inline]{Kyle:#1}}
    \newcommand{\LS}[1]{\todo[color=green,inline]{Hocka:#1}}
    \newcommand{\HP}[1]{\todo[color=green,inline]{HP:#1}}
    \newcommand{\PP}[1]{\todo[color=lime,inline]{Parth: #1}}
    \newcommand{\JL}[1]{\todo[color=cyan,inline]{Josh:#1}}
    \newcommand{\WJ}[1]{\todo[color=pink,inline]{Wenxin: #1}}
    \newcommand{\JS}[1]{\todo[color=blue,inline]{Josh: #1}}
    \newcommand{\HH}[1]{\todo[color=orange,inline]{Hyeonwoo: #1}}
    \newcommand{\KC}[1]{\todo[color=cyan,inline]{Kelechi: #1}}
    
\else
    \newcommand{\AH}[1]{}
    \newcommand{\AM}[1]{}
    \newcommand{\JD}[1]{}
    \newcommand{\TL}[1]{}
    \newcommand{\PA}[1]{}
    \newcommand{\KR}[1]{}
    \newcommand{\LS}[1]{}
    \newcommand{\HP}[1]{}
    \newcommand{\PP}[1]{}
    \newcommand{\JL}[1]{}
    \newcommand{\WJ}[1]{}
    \newcommand{\JS}[1]{}
    \newcommand{\HH}[1]{}
    \newcommand{\KC}[1]{}
    
\fi

\usepackage{hyperref}

\usepackage{cleveref}
\crefformat{section}{\S#2#1#3}
\crefname{figure}{Figure}{Figures}
\crefname{table}{Table}{Tables}
\crefname{theorem}{Theorem}{Theorems}
\crefname{thm}{Theorem}{Theorems}
\crefname{lemma}{Lemma}{Lemmata}
\crefname{equation}{Eqt.}{Eqts.}
\crefformat{Grammar}{Grammar #1}
\crefname{appendix}{Appendix}{Appendices}
\crefname{listing}{Listing}{Listings}

\usepackage{url}

\usepackage{tcolorbox}

\makeatletter
\newcommand{\linebreakand}{%
  \end{@IEEEauthorhalign}
  \hfill\mbox{}\par
  \mbox{}\hfill\begin{@IEEEauthorhalign}
}
\makeatother

\usepackage{pifont}

 \usepackage{pifont}
\begin{document}

\title{Recommending Pre-Trained Models for IoT Devices}

\ifANONYMOUS
\author{
{\rm Anonymous author(s)}
}
\else


\author{
    \IEEEauthorblockN{Parth V. Patil, Wenxin Jiang, Huiyun Peng,\\ Daniel Lugo, Kelechi G. Kalu}
    \IEEEauthorblockA{\textit{Purdue University}}
    \and
    \IEEEauthorblockN{Josh LeBlanc, Lawrence Smith,\\ Hyeonwoo Heo, Nathanael Aou}
    \IEEEauthorblockA{\textit{Purdue University}}
    \and
    \IEEEauthorblockN{James C. Davis}
    \IEEEauthorblockA{\textit{Purdue University}}
}

\fi

\maketitle

\begin{abstract}

The availability of pre-trained models (PTMs) has enabled faster deployment of machine learning across applications by reducing the need for extensive training. Techniques like quantization and distillation have further expanded PTM applicability to resource-constrained IoT hardware. Given the many PTM options for any given task, engineers often find it too costly to evaluate each model’s suitability.  Approaches such as LogME, LEEP, and ModelSpider help streamline model selection by estimating task relevance without exhaustive tuning. However, these methods largely leave hardware constraints as future work—a significant limitation in IoT settings. In this paper, we identify the limitations of current model recommendation approaches regarding hardware constraints and introduce a novel, hardware-aware method for PTM selection. We also propose a research agenda to guide the development of effective, hardware-conscious model recommendation systems for IoT applications.


\end{abstract}


\section{Introduction}
Many IoT applications today use deep neural networks (DNNs) for advanced functionality. Examples include object detection in autonomous vehicles \cite{shah_survey_2022}, crop health monitoring in agriculture \cite{pratyush_reddy_iot_2020}, and natural language processing for smart home assistants \cite{rani_voice_2017_no_url}. Given the cost of developing and training a DNN from scratch \cite{almeida_smart_2021}, engineers often leverage pre-trained models (PTMs) to expedite development and deployment processes. However, as highlighted by recent studies \cite{jiang_empirical_2023}, selecting an appropriate PTM frequently requires manual evaluation of model performance on downstream tasks, a process that is time-intensive and prone to variability. Furthermore, a separate study \cite{gopalakrishna_if_2022} found that 
practitioners often encounter hardware constraints or lack the expertise needed to adapt DNN as the main hurdle. This underscores the value of a systematic approach to PTM recommendation, particularly one that addresses the resource limitations of IoT hardware.

\begin{figure}[t]
    \centering
    \includegraphics[width=\linewidth, trim={1.5cm 9cm 1.5cm 1.5cm}, clip]{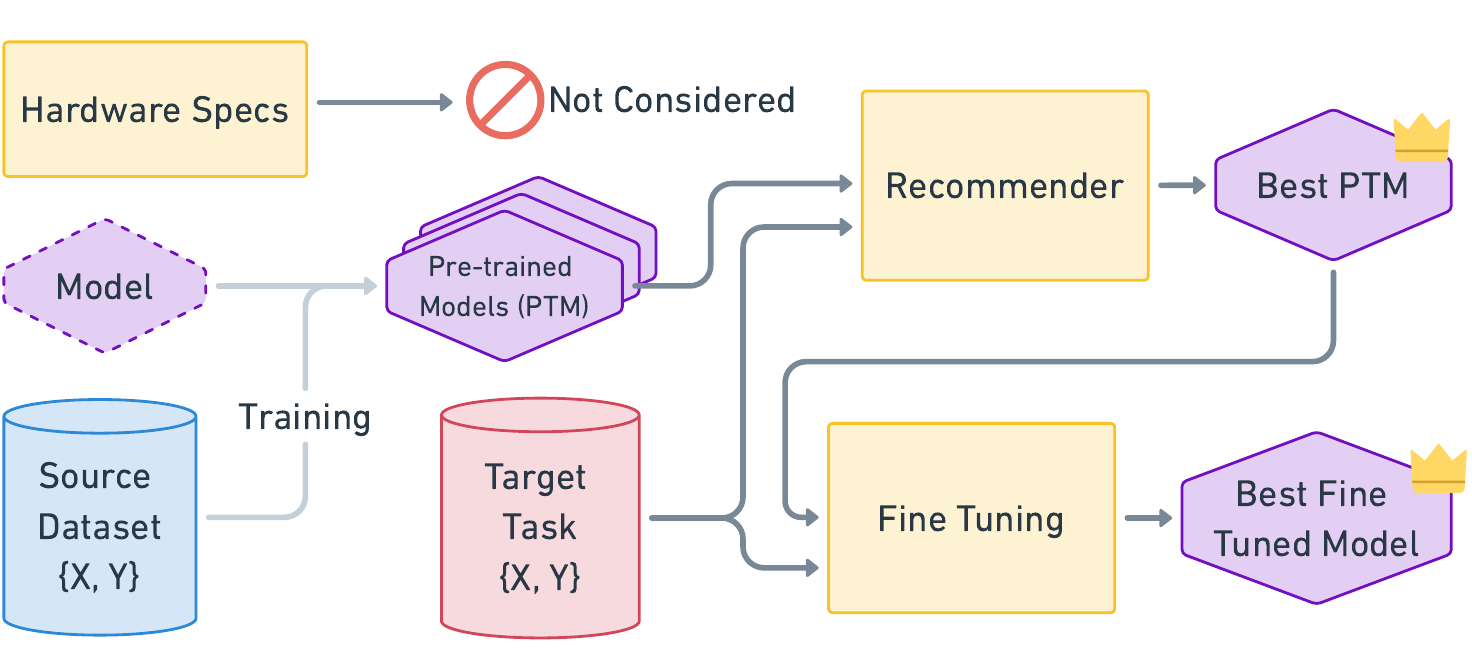}
    \caption{
    A model recommendation process for selecting the best pre-trained model (PTM) for a target task. Notably, hardware specifications are not considered, limiting use on constrained devices. Models trained on a source dataset (X, Y) are stored in a model hub. For a target task with a non-overlapping dataset (X, Y), the system recommends the most suitable PTM for fine-tuning, resulting in the best model for the task.}
    \vspace{-15pt}
    \label{fig:intro}
\end{figure}

PTM selection extends beyond IoT, with prior work making significant strides in recommending PTMs for specific tasks without model fine-tuning. Figure \ref{fig:intro} summarizes the current PTM selection pipeline. These methods fall into two categories: heuristic-based \cite{tran_transferability_2019, bao_information-theoretic_2019, nguyen_leep_2020, li_ranking_2021, you_logme_2021, ding_pactran_2022, pandy_transferability_2022, deshpande_linearized_2021} and learning-based \cite{zhang_model_2023, meng_foundation_2023, bai_pre-trained_2024}, as detailed in \ref{background}. However, their primary focus is maximizing model performance (\eg accuracy), 
with limited attention to hardware constraints. Deploying PTMs on IoT devices introduces many such constraints, including limited CPU, memory, energy, and low-bandwidth communication.
Adapting current recommendation approaches to address these IoT-specific limitations is nontrivial, as many of these methods rely on measures such as class similarity between source and target tasks or the model’s capacity for distinguishing classes.
Therefore, an effective IoT solution must incorporate both task suitability and hardware constraints, with potential further insights from modeling device energy consumption.
We identify two critical gaps in the state-of-the-art PTM recommendation systems:
  (1) \textit{Lack of IoT-specific inputs for model recommendation} 
  (2) \textit{Lack of Ground-Truth rankings for IoT devices}.
  Developing a hardware-aware recommender requires first establishing a ground truth model ranking on devices, which is essential for addressing the
first gap.

In this work, we introduce approaches and research agendas to address these issues.
 First, we propose two modifications to the Model Spider framework \cite{zhang_model_2023} to enable hardware-aware recommendations, which we term \textit{Model Spider Fusion} and \textit{Model Spider Shadow}. Both approaches aim to address the first gap, each with distinct methods. Additionally, we outline methods for collecting raw data by defining essential metrics and generating custom rankings based on these hardware-specific performance indicators.
We close by discussing future opportunities to support the ongoing development of hardware-conscious model recommendation systems.

Our contributions are:
\begin{enumerate}
    \item We identify key gaps in PTM recommendation for IoT.
    \item  We introduce methods to address these gaps, focusing on hardware-specific metrics and tweaks to existing methods.
    \item We outline a research agenda aimed at advancing PTM recommendations with a focus on hardware and sustainability.
\end{enumerate}


\section{background and related work}
\label{background}
This section covers background and related works on pre-trained model recommendation (\cref{sec:background-modelRecommendation}) and model benchmarking in IoT systems (\cref{sec:background-IoTBenchmarking}).

\subsection{Works for Model Recommendation}
\label{sec:background-modelRecommendation}
Researchers have developed various methods to help software engineers identify the most suitable PTMs, aiming to minimize fine-tuning and forward passes to reduce computational costs, as summarized in \cref{tab:comparison_rec_framework}.
This section reviews state-of-the-art approaches for PTM recommendations in downstream tasks, categorizing them into two main types: \textit{heuristic-based} and \textit{learning-based}.

\textbf{Heuristic-Based Methods}: These methods typically devise a novel heuristic or scoring system to rank a PTM's effectiveness for a target task. 
This category includes methods, such as NCE \cite{tran_transferability_2019}, and H-Score \cite{bao_information-theoretic_2019}, which estimate the similarity between source and target labels directly. 
Additional methods, such as LEEP \cite{nguyen_leep_2020}, N-LEEP \cite{li_ranking_2021}, LogME \cite{you_logme_2021}, PACTran \cite{ding_pactran_2022}, GBC \cite{pandy_transferability_2022}, and LFC \cite{deshpande_linearized_2021}, rely on a forward pass on the target dataset to capture representations. These representations are then analyzed for their usefulness based on target labels.
However, heuristic approaches often struggle to capture the nuances of hardware specifications and their correlation with model performance, overlooking the practical challenges software engineers face when deploying models on hardware-constrained environments.
For example, on IoT devices, larger models can leverage hardware-accelerated activation functions to boost performance, while smaller models with complex functions may still encounter bottlenecks.

\textbf{Learning-Based Methods}: Recent methods, including Model Spider \cite{zhang_model_2023},  EMMS \cite{meng_foundation_2023}, and Fennec \cite{bai_pre-trained_2024}, employ machine learning to eliminate the need for forward passes. These methods encode both the PTM and target dataset into a latent space, using learning techniques to discern patterns and predict performance. Given their promising results, we focus on learning-based techniques for our hardware-aware recommendation solutions, as this approach is well-suited for software engineers to encode complex hardware specifications, such as CPU type, architecture, memory size, and I/O speed.

Model Spider tokenizes tasks and pre-trained models to generate recommendations that balance efficiency and accuracy~\cite{zhang_model_2023}.
The Model Spider approach encodes each pre-trained model into a token $ \theta_m $ using an extractor $ \Psi $, which encapsulates essential characteristics such as architectures and parameters. Simultaneously, tasks are embedded as other tokens $\mu(T)$, reflecting dataset statistics and task characteristics. The key contribution of Model Spider is its use of a multi-head attention mechanism to assess the similarity $ sim(\theta_m, \mu(T)) $ between model and task tokens. This similarity score serves as a reliable indicator of model performance, enabling a ranked selection of pre-trained models aligned with the task requirements. 
Optionally, it also allows for the application of a forward pass, leading to a refined token $\theta^*_m$ that further captures data-specific attributes. This is done for 
top $K$-ranked models from the first iteration.
Unlike EMMS \cite{meng_foundation_2023} and Fennec \cite{bai_pre-trained_2024}, Model Spider’s tokenized, attention-based design is more suitable to capture the nuances of hardware specifications.
\WJ{^Be more specific, why it's more suitable. Because it's easy to add more features?}

\begin{table}[H]

    \centering
    
    \caption{This table compares current model recommendation approaches and their requirements. ModelSpider is notably efficient, not requiring a forward pass on the target dataset; however, none address hardware constraints—a key limitation for resource-constrained IoT environments.}
    \label{tab:comparison_rec_framework}
    \begin{tabular}{c|c|c|c|c}
    \toprule
    \rowcolor[HTML]{EFEFEF}

    \multicolumn{1}{c|}{\cellcolor[HTML]{EFEFEF}\textbf{Recommendation }} & \multicolumn{3}{c|}{\cellcolor[HTML]{EFEFEF}\textbf{Requires}} &  \multicolumn{1}{c}{\cellcolor[HTML]{EFEFEF}\textbf{Hardware}} \\

    \hhline{|~|---|~|}
    \rowcolor[HTML]{EFEFEF} 
    
    \multicolumn{1}{c|}{\cellcolor[HTML]{EFEFEF}\textbf{Approach }} & 
    \multicolumn{1}{c|}{\cellcolor[HTML]{EFEFEF}\textit{Forward }} & 
    \multicolumn{1}{c|}{\cellcolor[HTML]{EFEFEF}\textit{Source }}& 
    \multicolumn{1}{c|}{\cellcolor[HTML]{EFEFEF}\textit{Target }}& 
    \multicolumn{1}{c}{\cellcolor[HTML]{EFEFEF} \textbf{Aware}} \\

     \rowcolor[HTML]{EFEFEF} 
    
     & 
    \multicolumn{1}{c|}{\cellcolor[HTML]{EFEFEF}\textit{ Pass}} & 
    \multicolumn{1}{c|}{\cellcolor[HTML]{EFEFEF}\textit{ Labels}}& 
    \multicolumn{1}{c|}{\cellcolor[HTML]{EFEFEF}\textit{ Labels}}& 
    \multicolumn{1}{c}{\cellcolor[HTML]{EFEFEF}} \\
    \hline
    \rowcolor[HTML]{F3F3F3}
    NCE \cite{tran_transferability_2019} & No & - & -  & $ \times $ \\
     H-Score \cite{bao_information-theoretic_2019} & No & - & - & $ \times $ \\
   \rowcolor[HTML]{F3F3F3}
     OTCE \cite{tan_otce_2021} & - & - & -  & $ \times $ \\ 
     LEEP \cite{nguyen_leep_2020} & - & No & -  & $ \times $ \\
    \rowcolor[HTML]{F3F3F3}
     N-LEEP \cite{li_ranking_2021} & - & No & -  & $ \times $ \\
     LogME \cite{you_logme_2021} & - & - & No  & $ \times $ \\
    \rowcolor[HTML]{F3F3F3}
     PACTran \cite{ding_pactran_2022} & - & - & - & $ \times $ \\
     GBC \cite{pandy_transferability_2022} & - & No & - & $ \times $ \\
     \rowcolor[HTML]{F3F3F3}
     LFC \cite{deshpande_linearized_2021} & - & No & - & $ \times $ \\
    \hline
     & & & & \\ 
     \rowcolor[HTML]{F3F3F3}
     Model Spider \cite{zhang_model_2023} & No & No & No & $ \times $ \\
     EMMS \cite{meng_foundation_2023} & No & No & - & $ \times $ \\
     \rowcolor[HTML]{F3F3F3}
     Fennec \cite{bai_pre-trained_2024} & No & No & - & $ \times $ \\
    \hline
     & & & & \\ 
     \rowcolor[HTML]{F3F3F3}
     MS Fusion  & No & No & No & Yes \\
     MS Shadow & No & No & No & Yes \\
     
    \bottomrule
\end{tabular}
\end{table}

\subsection{Works for Model Benchmarking on IoT}
\label{sec:background-IoTBenchmarking}

Several studies have focused on benchmarking machine learning model performance on IoT devices, addressing specific applications and sustainability considerations relevant to software engineering workflows. For example, LwHBench \cite{sanchez_sanchez_lwhbench_2023} provides performance benchmarks to assess model performance, they use it for applications in agriculture, while Sayeedi et al. \cite{sayeedi_comparative_2024} emphasize the sustainability of models in terms of environmental impact. However, these approaches primarily evaluate models individually rather than comparing and ranking them. Sayeedi et al. introduce innovative metrics such as the Green Carbon Footprint, which combines power consumption, execution time, and other factors to provide a more holistic assessment of a model’s environmental impact. Our work extends these studies by incorporating these metrics to provide a systematic way for software engineers to rank model performance across IoT devices, facilitating more informed and environmentally conscious model recommendations.





\begin{figure*}[t]
    \centering
    \includegraphics[width=\linewidth, trim={1cm 6.5cm 1cm 6.5cm}, clip]{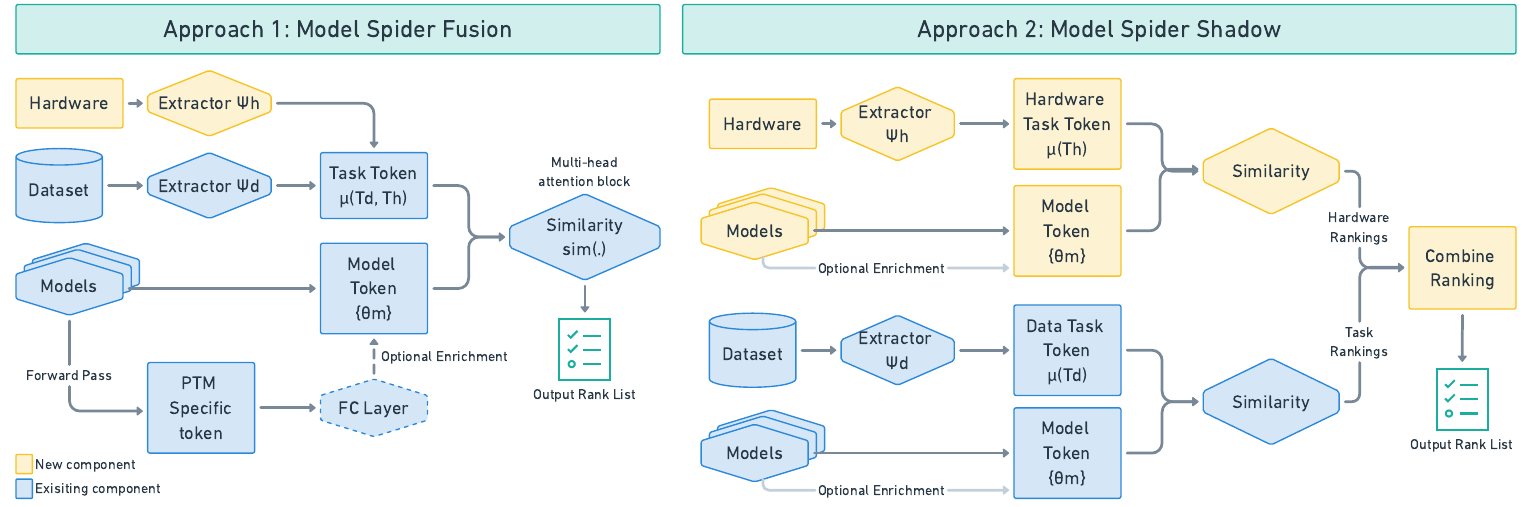}
    \caption{
     Overview of our proposed IoT-specific model recommendation approaches.
     Yellow components indicate new modules introduced, and blue represents existing Model Spider components.
     \textit{Model Spider Fusion} incorporates hardware specifications directly into task tokens via a hardware extractor. \textit{Model Spider Shadow} creates dual ranking systems—task relevance and hardware compatibility—combined through Copeland’s method for balanced recommendations. 
    }
    \label{fig:model_spider_approaches}

    \vspace{-15pt}
\end{figure*}

\section{Motivation and Gap Analysis}
While previous work has focused on model recommendations for downstream reuse, no studies or data currently support hardware-aware model selection. This section outlines key gaps in hardware awareness (\cref{sec:motivation-HWAwareness}) and the lack of ground truth (\cref{sec:motivation-GTDataset}) in model recommender works.


\subsection{Lack of IoT specific inputs for model recommendation}
\label{sec:motivation-HWAwareness}
\textbf{A key limitation of current model recommendation approaches is that they fail to consider the hardware performance implications while selecting models.}
For example, a recommendation system may rank models M1, M2, and M3, prioritizing M1 based on task performance. However, if M1 is a large model requiring extensive time—such as 10 minutes for a forward pass on certain IoT hardware—it may be impractical for real-time applications, making M2 the more effective choice by balancing accuracy with practical performance constraints. Furthermore, deep learning compilers like ONNX
could also influence model rankings by optimizing specific models for certain hardware types\cite{davis_reusing_2023, jajal2024interoperability}.
Nevertheless, performance remains a major challenge for engineers when reusing PTMs, as highlighted by \cite{jiang_challenges_2023}, and this issue is particularly critical in IoT contexts where resource limitations amplify its impact. As one participant in that study noted, ``\textit{Performance bugs are silent bugs. You will never know what happened until it goes to deployment.}'' This emphasizes the need for a structured approach to assess a model’s overall performance on IoT devices.

To further explore the identified research gaps, we propose the following research questions:

\todo[color=lightcyan,inline]{\textbf{RQ1}: How can different parameters be weighted to tailor recommendations? Specifically, can the solution be tuned to prioritize speed, cost-efficiency, or energy efficiency?\\\textbf{RQ2}: How do model optimization techniques, such as quantization \cite{jacob_quantization_2018} and distillation \cite{liu_improving_2018}, impact model rankings in the context of hardware-aware recommendations?}
\vspace{-5pt}

\subsection{Lack of Ground-Truth rankings for IoT devices}
\label{sec:motivation-GTDataset}

A fundamental barrier to progress in the earlier gap is the \textbf{lack of empirical data comparing the performance of different models across various IoT devices}. 
This gap prevents researchers from finding patterns or correlations in model parameters that could provide a quick, reliable way to assess hardware compatibility.
Previous benchmarking works \cite{sayeedi_comparative_2024, sanchez_sanchez_lwhbench_2023}, only consider one model and do not compare or rank them. 
The lack of this data restricts our ability to develop predictive models to recognize patterns in rankings. Model Spider \cite{zhang_model_2023} includes a method to approximate ground truth by combining multiple approaches using Copeland's rank aggregation \cite{copeland1951, saari_copeland_1996}. However, since there are few heuristic ranking approaches in the hardware space, this strategy is currently infeasible. This lack of comprehensive performance data across the hardware landscape highlights the need for dedicated datasets and benchmarks to enable informed and hardware-aware model recommendations.

The following research questions will guide our analysis after data collection, aiming to uncover trends and correlations:

\todo[color=lightcyan,inline]{\textbf{RQ3}: Can specific trends be identified between hardware specifications and model performance?\\\textbf{RQ4}: Is there a significant correlation between hardware resources and model characteristics, such as parameters, architecture, or problem type?}

\section{Research Agenda}

This section outlines a research agenda to enhance model recommendations for IoT applications. We introduce two approaches to make Model Spider hardware-aware (\cref{sec:NextStep-ModelSpiderModification}) and propose methods for dataset creation (\cref{sec:NextStep-Dataset}) , improving adaptability across IoT environments. We then discuss extending our approach to broader deep learning systems and complex model reuse scenarios (\cref{sec:FutureDirections}).

\subsection{Modifications to Model Spider}
\label{sec:NextStep-ModelSpiderModification}

To address the first gap identified, we propose two approaches to make the existing Model Spider framework hardware-aware. These build upon the existing framework, leveraging its robustness while enhancing its capability to account for hardware constraints with minimal modifications.


\subsubsection{\textbf{Model Spider Fusion -- Augmenting Task Tokens}}
\label{next_steps_MS_Fusion}
        As shown in Figure \ref{fig:model_spider_approaches}, in this approach we propose to introduce a separate Extractor $ \Psi_h $ which would encode the hardware specification for any given hardware. Its inputs comprise hardware model, CPU specifications, RAM size, Memory Size, etc. 
        We would append these to the existing Task Tokens $ \mu(T_d, T_h) $. This modification enables the similarity block to learn correlations between model performance and the specific hardware in use, thereby enhancing the hardware-awareness.
    
\subsubsection{\textbf{Model Spider Shadow -- New Hardware task}}
\label{next_steps_MS_Shadow}


        This approach builds on Model Spider's capability to recommend the best model for a downstream task, expanding it to include hardware requirements. By redefining a ``downstream task'' to encompass specific hardware, we replicate the framework with a hardware extractor $ \Psi_h $ that encodes hardware-specific features. This results in two ranking systems: a task-based selector for model suitability to tasks and a hardware-based selector for compatibility with hardware. We combine these rankings using Copeland’s method \cite{copeland1951, saari_copeland_1996}, yielding a balanced recommendation that accounts for both task and hardware. This framework can be further extended to include energy-based or cost-based selectors, enabling tailored recommendations.

\subsection{Dataset Creation}
\label{sec:NextStep-Dataset}

\textbf{Defining Metrics:} 
 We propose to use metrics laid out in Sayeedi \etal \cite{sayeedi_comparative_2024}, categorized into two groups. These groups can either be combined to establish the ground truth in \textit{Model Spider Fusion} (\cref{next_steps_MS_Fusion}) or used independently to rank models in \textit{Model Spider Shadow} (\cref{next_steps_MS_Shadow}).

\begin{table}[t]
    \centering
    \caption{
    Categorizing Metrics into Groups
    }
    \label{tab:performance_metrics}
    \begin{NiceTabular}{lllc}
        \toprule
         & \textbf{Metric} & \textbf{Description} & \textbf{Sample}\\        
        \hline
        \Block{6-1}{\rotate Hardware} \\
        \rowcolor[HTML]{F3F3F3}
         & Execution Time & Time for one forward pass & 90 ms \\
         & Memory Utilization & Memory used during execution & 3 GB \\
         \rowcolor[HTML]{F3F3F3}
         & Power Consumption & Total energy consumed & 5 W \\
         & CPU Temperature & Temperature of the CPU & 75°C \\
         \rowcolor[HTML]{F3F3F3}
         &  \Block{2-1}{Carbon \\Footprint~\cite{sayeedi_comparative_2024}} & \Block{2-1}{Environmental impact}  & 10.9 \\
         \rowcolor[HTML]{F3F3F3}
         & & & units\\
         \midrule
        \Block{4-1}{\rotate Model} \\
         & Accuracy & \% correct predictions & 92\% \\
         \rowcolor[HTML]{F3F3F3}
         & Precision & identified / predicted positives  & 88\% \\
         & Recall (Sensitivity) & identified / actual positives & 85\% \\
         \rowcolor[HTML]{F3F3F3}
         & F1 Score & Harmonic mean of last two  & 86\% \\
         \bottomrule
    \end{NiceTabular}
    \vspace{-10pt}
\end{table}



\paragraph{Methodology to collect data}
Additionally, to construct the ground truth data, we consider multiple PTMs and datasets. Each PTM is fine-tuned on all datasets. A robust methodology is then defined for collecting data from these (Fine-Tuned Model, Dataset) pairs, ensuring that experiments do not interfere with previous runs and that sensitive metrics, such as CPU temperature, are accurately reported.

\paragraph{Creating rankings from collected data}

We propose a tunable ranking system that adjusts to specific requirements, ranking models based on selected metrics such as best execution time, best accuracy, or best energy consumption. To aggregate these metrics, we employ the weighted Copeland's rank-choice voting method. Each metric rank is assigned a weight $ w_i $ in $ [0,1] $, with $ \sum w_i = 1 $. The objective function \eqref{eq:objective} maximizes each model's combined score, incorporating both model performance and hardware efficiency \cite{Benmeziane2021ACS, Tan2019}. Here, \( f(\alpha) \) denotes model performance for configuration \( \alpha \) (\eg based on accuracy), and \( \text{HW}_i(\alpha) \) represents the \( i \)-th hardware metric (\eg execution time or power consumption). Each hardware metric is normalized by a threshold \( T_i \) and scaled by adjustable \( w_i \) to reflect the relevant importance.

\begin{equation}
\max_{\alpha \in A} f(\alpha) \cdot \sum_{i} \left( \frac{\text{HW}_i(\alpha)}{T_i} \right)^{w_i} \tag{1} \label{eq:objective}
\end{equation}

\subsection{Future Directions}
\label{sec:FutureDirections}
We propose to extend the model recommender to a broader deep learning system (\cref{sec:FutureDirection-BroaderDLSystem}) and complex reuse (\cref{sec:FutureDirection-ComplexModelReuse}).

\subsubsection{Extending to Broader Deep Learning Systems}
\label{sec:FutureDirection-BroaderDLSystem}
Our proposed approach would extend Model Spider, which handles only basic tasks (\eg classification) on edge devices. 
However, hardware constraints are a crucial engineering consideration not only on edge devices but across a range of deep learning systems, including distributed learning frameworks (\eg federated learning \cite{mcmahan_communication-efficient_2017}). Additionally, in real-world applications, the interaction between data collection and model prediction components must be addressed~\cite{makhshari2021IoTBugStudy, garcia2020AVBugs}.
We propose broadening our research to include DL systems, enhancing model reuse and adaptability across diverse environments.

\subsubsection{Extending to More Complex Model Reuse Scenarios}
\label{sec:FutureDirection-ComplexModelReuse}
Our work, along with most prior ML research, has primarily addressed basic tasks, such as classification and regression. 
Although Model Spider has explored model reuse in generative models (\eg Large Language Model)~\cite{zhang_model_2023}, broader applications of DL model reuse remain largely unexplored. 
In computer vision, for instance, classification serves as a foundation for more complex tasks, like object detection and segmentation, which build upon classification models as 
backbones~\cite{banna2021experience}. These downstream tasks require fine-tuning the model and adding task-specific heads or decoders for complex objectives. Expanding our approach to support such reuse scenarios is essential.

\definecolor{codegray}{rgb}{0.5,0.5,0.5}
\definecolor{darkgreen}{rgb}{0,0.5,0}
\definecolor{darkred}{rgb}{0.7,0,0}
\definecolor{purple}{rgb}{0.5,0,0.5}
\definecolor{blue}{rgb}{0,0,0.5}
\definecolor{magenta}{rgb}{0.5,0,0.5}
\lstdefinestyle{mystyle}{
  numberstyle=\tiny\color{codegray},
  basicstyle=\ttfamily\footnotesize,
  keywordstyle=\color{black},
  breakatwhitespace=false,         
  breaklines=false,                 
  captionpos=t,                    
  keepspaces=true,                 
  numbers=left,                    
  numbersep=3pt,                  
  showspaces=false,                
  showstringspaces=false,
  showtabs=false,                  
  tabsize=10,
  xleftmargin=10pt, 
  numbersep=5pt,  
  frame= tb,
  title={\vspace{-8pt}\hrulefill\vspace{5pt}\parbox[t]{\linewidth}{\textbf{   Algorithm 1} Performance Evaluation Procedure for\\
  \textbf{\bigskip} (Fine-Tuned Model, Dataset) Pairs}\vspace{-8pt}\hrulefill}
}
\lstset{style=mystyle}
\begin{lstlisting}[escapechar=|]
|\textbf{\color{blue}Input}:| Datasets and Models Fine-tuned on those
|\textbf{\color{blue}Output}:| Metrics contained in Table II
|\textbf{\color{blue}Function}| |\color{purple}stabilize()|:
  |\color{darkgreen}Wait| until:
    1) |\color{purple}CPU| and |\color{purple}RAM| utilization at nominal level
    2) |\color{purple}Temperature| within specified range
|\textbf{\color{blue}Function}| |\color{purple}benchmark(Datasets, Models)|:
  |\color{darkgreen}Call| stabilize()

  |\textbf{\color{magenta}while}| dataset, model in pairs(Datasets, Models)
    |\textbf{\color{magenta}for}| batch_size in range(1, 101)
      |\color{darkgreen}Sample| batch with current batch_size
      |\color{darkgreen}Perform| forward pass
      |\color{darkgreen}Measure| performance metrics

    |\color{darkgreen}Set| batch_size = |\color{purple}32|
    |\textbf{\color{magenta}for}| each batch in dataset
      |\color{darkgreen}Sample| batch
      |\color{darkgreen}Perform| forward pass
      |\color{darkgreen}Measure| performance metrics

    |\color{darkgreen}Call| stabilize()
\end{lstlisting}

\pagebreak

\section{Conclusion}
In conclusion, this paper identifies and addresses critical gaps in PTM recommendation for IoT by proposing hardware-aware enhancements to the Model Spider framework. This work establishes a foundation for further research in optimizing PTM recommendations across diverse IoT environments.

\bibliographystyle{IEEEtran}
\bibliography{refs/references, refs/new, refs/IoT_References}

\end{document}